\documentclass{article}
\usepackage{spconf,amsmath,graphicx,booktabs,array,url}
\usepackage{fontspec}
\usepackage{xeCJK}
\usepackage[hidelinks]{hyperref}

\setCJKmonofont{FandolSong-Regular.otf}
\newcommand{\target}[1]{\begingroup\xeCJKsetup{CJKecglue={}}\mbox{#1}\endgroup}
\newcommand{\targetdot}{\ensuremath{\cdot}}

\title{CN-NEWSTTS BENCH: A TARGET-LEVEL AUTOMATIC BENCHMARK FOR RAW-INPUT CHINESE NEWS TTS PRONUNCIATION}
\name{Shijun Luo\sthanks{First author and corresponding author.}}
\address{NetEase Cloud Music, Hangzhou, China}

\begin{document}
\ninept
\raggedbottom
\maketitle

\begin{abstract}
Chinese news text contains dense written forms such as scores, hyphenated
model names, ranges, unit symbols, percentages, English abbreviations, and
mixed Chinese--Latin--digit names. These forms are frequent in real listening
workflows, and a text-to-speech (TTS) system can preserve the written string
while changing the spoken meaning. We introduce CN-NewsTTS Bench v0.1, an open
target-level benchmark for evaluating whether Chinese news TTS products
pronounce such targets correctly from raw text, without user-side rules, LLM
rewriting, SSML hints, or manual edits. The release contains a 200-record
development set, an 800-record public test set, 992 public auto-evaluable
targets, fixed transcripts from a three-ASR ensemble, an automatic target
scorer, and initial results for seven product TTS systems. We additionally
report ASR-route diagnostics, ASR-subset ablations, category-level results,
confidence intervals, and provider configuration metadata. The best system
reaches 0.879 strict accuracy, while several systems remain below 0.60.
\end{abstract}


\section{Introduction}
\label{sec:intro}

Modern Chinese news TTS systems are used in settings where listeners expect
symbols and compact written forms to be read according to news conventions. A
short article may contain a sports score such as \target{96-91}, an aircraft
model such as \target{苏-27}, a year range such as \target{2028-2030年}, a torque
unit such as \target{620N\targetdot{}m}, or a generation label such as
\target{80后}. These strings are usually unambiguous to a human editor, yet the
same surface marks can represent ranges, subtraction, model designators, unit
names, or ordinary numbers.

Such failures are not merely naturalness problems. Reading a score as a range
changes the relation between the numbers; reading \target{苏-27} as a negative
number changes a military model into a nonsensical phrase; reading
\target{80后} as \target{八十后} loses the conventional generation-label
reading. Subjective tests such as MOS remain important for speech quality
\cite{itu_p800}, and objective predictors can estimate global naturalness or
voice-conversion quality \cite{lo2019mosnet}. They are, however, too coarse to
track specific high-impact reading decisions. Generic ASR round-trip
comparison is also insufficient because the evaluation must decide whether a
particular target was read as an intended news form.

The problem is closely related to text normalization and non-standard-word
handling in TTS \cite{taylor2009tts,sproat2001normalization,ebden2015kestrel,sproat2016rnn},
Mandarin text normalization \cite{zhou2008mandarin_tn,dai2022chinese_tn}, and
Chinese G2P/polyphone benchmarks \cite{park2020g2pm,zhang2023cvtepoly}.
CN-NewsTTS Bench differs by measuring the end-to-end raw-input behavior exposed
by deployed TTS APIs rather than an isolated frontend component. The main
contributions are:
\begin{itemize}
    \item an open, deterministic dev/public split for Chinese news pronunciation targets;
    \item a Raw Input Product Track that excludes user-side normalization or hints;
    \item a target-level schema with positive and negative readings;
    \item a three-ASR automatic scoring protocol with route diagnostics and ablations; and
    \item a reproducible v0.1 leaderboard for seven product TTS systems.
\end{itemize}

\section{Benchmark Design}
\label{sec:design}

\subsection{Raw Input Product Track}

The main track measures product-facing behavior under a fixed raw-input
condition. Each system receives the same original Chinese news-style text.
Provider-internal text normalization and frontend behavior are allowed, but
external rule frontends, LLM rewrites, SSML pronunciation hints, and manual
benchmark-text edits are not allowed. A fixed voice is chosen before
evaluation and kept constant for all samples. This is therefore a raw-input
product evaluation, not a claim that the benchmark isolates model weights or
removes all provider-side frontend logic.

\subsection{Data Construction and Targets}

The v0.1 data are deterministic synthetic news-style sentences generated from
category-specific templates and lexicons, not reused copyrighted news articles.
The released generator script uses seed 20260620, 11 template families, and
86 lexicon entries covering teams, events,
projects, cities, companies, military models, vehicle models, abbreviations,
mixed brand strings, unit symbols, generation labels, and polyphone audit
items. The templates mirror production risk motifs such as sports recaps,
defense/vehicle models, technical units, mixed brands, and time spans; they are
diagnostic rather than a claim about full news frequency. The dev and public
splits are constructed from fixed per-category quotas, then shuffled. Each
record contains targets with offsets, categories, positive/negative readings,
and an auto-evaluable flag. Optional same-character polyphones are included for
audit but excluded from the main automatic score because ordinary ASR text often
hides tonal distinctions.

\begin{center}
\refstepcounter{table}\label{tab:data}
\textbf{Table~\thetable.} CN-NewsTTS Bench v0.1 data scale.\\[-0.2em]
\begin{tabular}{lrrrr}
\toprule
Split & Rec. & Targets & Auto & Opt. \\
\midrule
dev & 200 & 252 & 248 & 4 \\
public & 800 & 1008 & 992 & 16 \\
total & 1000 & 1260 & 1240 & 20 \\
\bottomrule
\end{tabular}
\end{center}

\begin{table}[t]
\centering
\caption{Representative target annotations. Positive and negative patterns
are simplified examples; the released JSONL contains the full pattern lists.}
\label{tab:examples}
\scriptsize
\setlength{\tabcolsep}{2.7pt}
\begin{tabular}{llll}
\toprule
Span & Type & Positive & Negative \\
\midrule
\target{96-91} & score & \target{九十六比九十一} & \target{九十六到九十一} \\
\target{苏-27} & model & \target{苏二七}; \target{苏二十七} & \target{苏负二十七} \\
\target{80后} & generation & \target{八零后} & \target{八十后} \\
\target{620N\targetdot{}m} & unit & \target{六百二十牛米} & \target{六百二十恩点米} \\
\bottomrule
\end{tabular}
\end{table}

The public split contains 224 sports-score targets, 192 military-model
targets, 104 abbreviations, 96 unit-symbol targets, 80 hyphenated ranges, 80
year ranges, 64 mixed brand names, 64 vehicle models, 56 percentages, and 32
generation labels. The public test set is a transparent diagnostic split rather
than a long-term anti-overfitting competition set; a hidden split can be added
if leaderboard gaming becomes a practical concern.

\section{Automatic Evaluation}
\label{sec:eval}

\subsection{Three-ASR Protocol}

The v0.1 public leaderboard uses three heterogeneous ASR routes: MiMo API ASR,
SenseVoiceSmall, and Paraformer-zh. SenseVoiceSmall and Paraformer-zh are
open-source local recognizers from the FunAudioLLM/FunASR ecosystem
\cite{sensevoice,funasr}; MiMo API ASR provides an API-based route. The public
release fixes all three ASR transcript files, so the leaderboard can be
reproduced without rerunning commercial TTS APIs or ASR services. For seven TTS
systems and 800 public records, the canonical transcript set contains 16,800
successful ASR rows and no missing public samples.

\subsection{Target-Level Voting and Metrics}

For each ASR transcript and target, the scorer applies Unicode, case,
punctuation, and whitespace normalization. It then matches negative-reading
patterns before positive-reading patterns. A single ASR route returns
\texttt{wrong} if a negative reading is found, \texttt{correct} if a positive
reading is found, and \texttt{unknown} otherwise. Majority voting gives final
\texttt{correct} or \texttt{wrong} if at least two routes agree; all remaining
combinations become \texttt{unknown}.

The primary metric is strict auto accuracy,
\begin{equation}
\mathrm{StrictAcc} = \frac{\#\mathrm{correct}}{\#\mathrm{auto\mbox{-}evaluable\ targets}}.
\end{equation}
We also report coverage and resolved accuracy:
\begin{align}
\mathrm{Coverage} &=
\frac{\#\mathrm{correct}+\#\mathrm{wrong}}
{\#\mathrm{auto\mbox{-}evaluable\ targets}},\\
\mathrm{ResolvedAcc} &=
\frac{\#\mathrm{correct}}{\#\mathrm{correct}+\#\mathrm{wrong}}.
\end{align}
Keeping unknowns in the strict denominator prevents a system from appearing
accurate only on the subset that the ASR ensemble can resolve.

\subsection{ASR Protocol Diagnostics}

Table~\ref{tab:asrdiag} pools all 6,944 public target instances
(7 systems $\times$ 992 targets). The three individual routes have similar
strict accuracy but different coverage behavior. MiMo API ASR is conservative:
it has high resolved accuracy when it resolves a target, but many unknowns.
Paraformer-zh and SenseVoiceSmall resolve more targets. Pairwise tri-label
agreement is 0.959 between Paraformer-zh and SenseVoiceSmall, but 0.595 and
0.604 between MiMo API ASR and the two local routes, showing that the ensemble
is heterogeneous. Overall, 41.7\% of target instances have at least one route
disagreement, while only 0.7\% are complete correct/wrong/unknown ties.

\begin{table}[t]
\centering
\caption{ASR-route diagnostics over 6,944 public target instances. R-Strict
denotes strict accuracy computed from one ASR route before majority voting.}
\label{tab:asrdiag}
\setlength{\tabcolsep}{4.2pt}
\begin{tabular}{lrrrr}
\toprule
Route & R-Strict & Cov. & Res. & Unk. \\
\midrule
MiMo API ASR & .513 & .569 & .902 & .431 \\
Paraformer-zh & .543 & .748 & .725 & .252 \\
SenseVoiceSmall & .537 & .728 & .738 & .272 \\
Majority vote & .540 & .730 & .739 & .270 \\
\bottomrule
\end{tabular}
\end{table}

\begin{center}
\refstepcounter{table}\label{tab:asrablation}
\textbf{Table~\thetable.} ASR-subset ablation for strict accuracy. P, S, and M
denote Paraformer-zh, SenseVoiceSmall, and MiMo API ASR; two-route settings
require agreement.\\[-0.2em]
\scriptsize
\setlength{\tabcolsep}{2.6pt}
\begin{tabular}{lrrrr}
\toprule
System & Full3 & P+S & M+S & M+P \\
\midrule
Volcano & .879 & .870 & .575 & .570 \\
Azure & .756 & .750 & .504 & .510 \\
Google & .604 & .592 & .404 & .408 \\
MiniMax & .548 & .542 & .366 & .360 \\
Aliyun & .472 & .468 & .337 & .339 \\
MiMo & .275 & .274 & .225 & .226 \\
AWS & .244 & .230 & .192 & .200 \\
\bottomrule
\end{tabular}
\end{center}

The ASR ablation in Table~\ref{tab:asrablation} preserves the complete
seven-system rank order under all three two-route subsets. The local
Paraformer+SenseVoice route changes only 88 of 6,944 final target labels
(1.27\%) relative to the three-route majority, and all such changes turn
resolved labels into \texttt{unknown}. We use the three-route majority because
the two local routes are highly correlated, while MiMo API ASR adds an
independent route. The lower M+S and M+P scores are a two-route coverage effect:
both routes must resolve and agree, so conservative MiMo outputs become
\texttt{unknown}. Thus the two-route variants are stability checks, not the
leaderboard protocol.

\section{Initial Public Results}
\label{sec:results}

We evaluate seven product TTS systems: Volcano/Doubao TTS, Azure Speech TTS,
Google Cloud TTS, MiniMax TTS, Aliyun CosyVoice, MiMo TTS, and AWS Polly. Each
system generates one audio file per public test record, using one fixed voice
and provider-level default normalization. Generated audio is not redistributed
in the GitHub repository because commercial provider terms differ; fixed ASR
transcripts are released as the reproducible scoring artifact.

\begin{figure*}[!t]
\centering
\includegraphics[width=.72\textwidth]{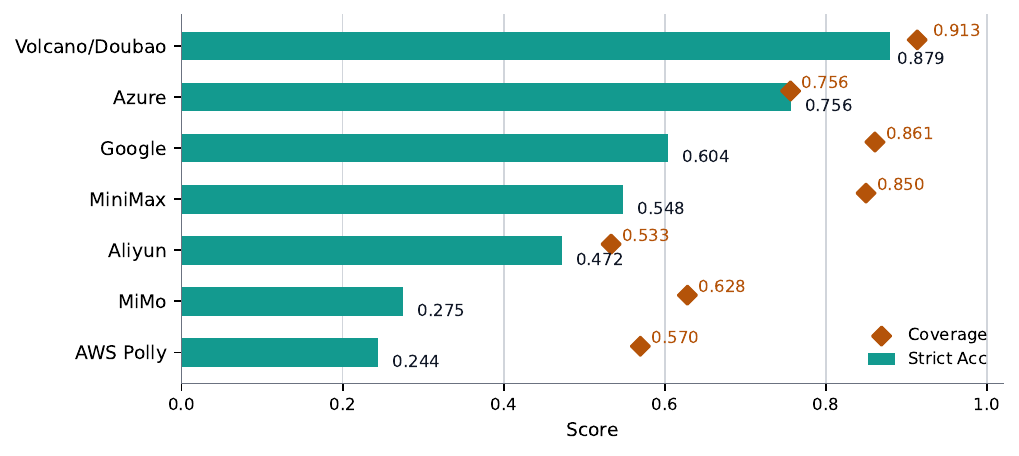}
\caption{Public test strict accuracy and coverage for seven raw-input TTS products.}
\label{fig:leaderboard}
\end{figure*}

\begin{table*}[!t]
\centering
\begin{minipage}[t]{0.48\textwidth}
\centering
\caption{Public test results on 992 auto-evaluable targets. Confidence
intervals are Wilson 95\% intervals for strict accuracy.}
\label{tab:leaderboard}
\scriptsize
\setlength{\tabcolsep}{2.0pt}
\begin{tabular}{lccccc}
\toprule
System & Strict & 95\% CI & Cov. & Res. & C/W/U \\
\midrule
Volcano & .879 & [.857,.898] & .913 & .962 & 872/34/86 \\
Azure & .756 & [.728,.782] & .756 & 1.000 & 750/0/242 \\
Google & .604 & [.573,.634] & .861 & .701 & 599/255/138 \\
MiniMax & .548 & [.517,.579] & .850 & .645 & 544/299/149 \\
Aliyun & .472 & [.441,.503] & .533 & .885 & 468/61/463 \\
MiMo & .275 & [.248,.304] & .628 & .438 & 273/350/369 \\
AWS & .244 & [.218,.272] & .570 & .428 & 242/323/427 \\
\bottomrule
\end{tabular}
\end{minipage}\hfill
\begin{minipage}[t]{0.43\textwidth}
\centering
\caption{Category-level public results pooled over the seven TTS systems.
The target count is per system; rates are over all systems.}
\label{tab:catresults}
\scriptsize
\setlength{\tabcolsep}{3.0pt}
\begin{tabular}{lrrrrr}
\toprule
Category & N & Strict & Cov. & Wrong & Unk. \\
\midrule
percentage & 56 & 1.000 & 1.000 & .000 & .000 \\
vehicle & 64 & .929 & .951 & .022 & .049 \\
abbrev. & 104 & .865 & .912 & .047 & .088 \\
range & 80 & .750 & .827 & .077 & .173 \\
year range & 80 & .713 & .713 & .000 & .288 \\
generation & 32 & .509 & .813 & .304 & .188 \\
military & 192 & .440 & .587 & .147 & .413 \\
brand & 64 & .422 & .578 & .156 & .422 \\
unit & 96 & .342 & .528 & .186 & .472 \\
sports & 224 & .233 & .728 & .494 & .272 \\
\bottomrule
\end{tabular}
\end{minipage}
\end{table*}

Figure~\ref{fig:leaderboard} and Table~\ref{tab:leaderboard} show substantial
variation. The best system reaches 0.879 strict accuracy, but several widely
used systems remain below 0.60. Azure has no majority-voted wrong targets, but
it has 242 unknown targets, so its strict accuracy is 0.756 rather than 1.0.
Google and MiniMax have higher coverage, but many more wrong targets. These
patterns show why strict accuracy, coverage, and resolved accuracy must be
reported together.

Table~\ref{tab:catresults} shows that the benchmark is not a single global
difficulty score. Percentages, vehicle models, and abbreviations are largely
solved by the evaluated systems. Sports scores are the hardest category by
strict accuracy and wrong rate: many systems read score hyphens as ranges or
subtractions. Unit symbols have the highest unknown rate, often because ASR
routes cannot reliably expose symbol-level readings.

The category averages hide system-specific failure modes, so
Table~\ref{tab:hardmatrix} breaks out the hardest categories by system.

\begin{center}
\refstepcounter{table}\label{tab:hardmatrix}
\textbf{Table~\thetable.} Per-system strict accuracy on selected hard categories.\\[-0.2em]
\centering
\scriptsize
\setlength{\tabcolsep}{2.8pt}
\begin{tabular}{lrrrrr}
\toprule
System & Sports & Unit & Mil. & Brand & Gen. \\
\midrule
Volcano & .996 & .573 & .865 & .391 & 1.000 \\
Azure & .504 & .646 & .771 & .438 & 1.000 \\
Google & .000 & .417 & .776 & .484 & .125 \\
MiniMax & .000 & .542 & .448 & .484 & .562 \\
Aliyun & .134 & .021 & .198 & .281 & .750 \\
MiMo & .000 & .198 & .000 & .438 & .000 \\
AWS & .000 & .000 & .026 & .438 & .125 \\
\bottomrule
\end{tabular}
\end{center}

Volcano nearly solves sports scores but still leaves units and mixed brands
difficult; Azure is conservative but strong on generation labels and units;
Google and MiniMax lose most sports-score targets even when they are competitive
elsewhere. For zero-score sports systems, the dominant failure is reading score
hyphens as ranges, e.g., \target{96-91} as \target{九十六到九十一}. Route-level
range negatives dominate subtraction negatives (1,496 vs. 266).

\section{Reproducibility and Release}
\label{sec:release}

\noindent{\footnotesize
Repository: \href{https://github.com/Jayden-X-L/cn-news-tts-bench}{\texttt{https://github.com/Jayden-X-L/}}\\
\href{https://github.com/Jayden-X-L/cn-news-tts-bench}{\texttt{cn-news-tts-bench}}\\
Release: \href{https://github.com/Jayden-X-L/cn-news-tts-bench/releases/tag/v0.1}{\texttt{https://github.com/Jayden-X-L/}}\\
\href{https://github.com/Jayden-X-L/cn-news-tts-bench/releases/tag/v0.1}{\texttt{cn-news-tts-bench/releases/tag/v0.1}}\\
Commit: \texttt{f94a679fc7fc}.}
The release includes data, schema, scoring code, fixed ASR transcripts,
leaderboard files, the dashboard, audit, and checksums. Table~\ref{tab:providers}
lists compact model and voice labels; region, sample rate, and invocation date
are recorded in the release metadata. The v0.1 data, TTS, and audit dates are
2026-06-20, 2026-06-22, and 2026-06-23.

\begin{center}
\refstepcounter{table}\label{tab:providers}
\textbf{Table~\thetable.} Initial TTS configuration.\\[-0.2em]
\centering
\scriptsize
\renewcommand{\arraystretch}{0.90}
\setlength{\tabcolsep}{2.0pt}
\begin{tabular}{lll}
\toprule
System & Model/API & Voice/config label \\
\midrule
Volcano & seed-tts-2.0-standard & default zh \\
Azure & azure-neural-tts & zh-CN-XiaoxiaoNeural \\
Google & Chirp 3 HD & cmn-CN-Chirp3-HD-Kore \\
MiniMax & speech-2.8-hd & Mandarin news \\
Aliyun & cosyvoice-v3-plus & longanyang \\
MiMo & mimo-v2.5-tts & \target{白桦} \\
AWS & polly-neural & Zhiyu \\
\bottomrule
\end{tabular}
\end{center}

A minimal reproduction consists of validating the public split, scoring a
per-model ASR result file, aggregating the leaderboard, and verifying the
checksum manifest. The three canonical ASR transcript files each contain 5,600
rows. New TTS systems can be evaluated by generating audio from the raw public
text and either running the provided ASR pipeline or submitting ASR result
files in the documented JSONL format.

Leaderboard entries should report strict accuracy, coverage, category
breakdowns, data release, provider, voice, region, sample rate, date, and ASR
files. Updated APIs should be added as dated entries rather than replacing the
baseline.

\pagebreak

\section{Limitations}
\label{sec:limitations}

The benchmark is automatic and should not replace human listening tests. ASR
text can hide pronunciation errors, especially same-character tones, and
unknown rates vary by category. The public split supports reproducibility but
may later require a hidden test split. Commercial TTS audio is omitted because
provider terms vary; the release fixes data, metadata, transcripts, scores, and
checksums for dated comparison.

\section{Conclusion}
\label{sec:conclusion}

CN-NewsTTS Bench v0.1 provides an open, target-level benchmark for raw-input
Chinese news TTS pronunciation accuracy. By fixing the data split, three-ASR
transcripts, scorer, category diagnostics, and initial seven-system
leaderboard, it makes a common production failure mode measurable and
reproducible. The results indicate that compact Chinese news forms remain a
practical challenge for current TTS products.

\section{Compliance with Ethical Standards}
This work uses synthetic news-style text and released benchmark transcripts.
No human-subject audio, personal speech data, or private user data is released.

\clearpage
\bibliographystyle{IEEEbib}
\bibliography{refs}

\end{document}